\documentclass[conference]{IEEEtran}
\IEEEoverridecommandlockouts
\usepackage{times}
\usepackage{epsfig}
\usepackage{graphicx}
\usepackage{amsmath}
\usepackage{amssymb}
\usepackage{multirow}
\usepackage{booktabs}
\usepackage{url}
\usepackage[table]{xcolor}
\usepackage{xcolor}
\usepackage{pgfplotstable}
\usepackage[left=0.75in, right=0.75in, top=1in, bottom=0.75in]{geometry}
\usepackage{url}
\usepackage{float}
\usepackage{arydshln}

\def\btex{B\kern-.05em{i}\kern-.025em{b}\kern-.08em\TeX}

\def\cca#1{%
	\pgfmathsetmacro\calc{100-(#1-0.1)*100/(1-0.1)}%
	\edef\clrmacro{\noexpand\cellcolor{violet!\calc}}%
	\clrmacro%
	\ifdim \calc pt>50pt\color{white}\fi{#1}%
}

\def\BibTeX{{\rm B\kern-.05em{\sc i\kern-.025em b}\kern-.08em
		T\kern-.1667em\lower.7ex\hbox{E}\kern-.125emX}}
\begin{document}
	
	\title{An Ensemble of Knowledge Sharing Models for Dynamic Hand Gesture Recognition}
	
	\author{\IEEEauthorblockN{Kenneth Lai and Svetlana Yanushkevich}
		\IEEEauthorblockA{Biometric Technologies Laboratory, Department of ECE, University of Calgary, Canada \\
			Email: {kelai, syanshk}@ucalgary.ca}
	}

	\maketitle
	
	\begin{abstract}
		
		The focus of this paper is dynamic gesture recognition in the context of the interaction between humans and machines. We propose a model consisting of two sub-networks, a transformer and an ordered-neuron long-short-term-memory (ON-LSTM) based recurrent neural network (RNN).  Each sub-network is trained to perform the task of gesture recognition using only skeleton joints.  Since each sub-network extracts different types of features due to the difference in architecture, the knowledge can be shared between the sub-networks.  Through knowledge distillation, the features and predictions from each sub-network are fused together into a new fusion classifier.  In addition, a cyclical learning rate can be used to generate a series of models that are combined in an ensemble, in order to yield a more generalizable prediction.  The proposed ensemble of knowledge-sharing models exhibits an overall accuracy of 86.11\% using only skeleton information, as tested using the Dynamic Hand Gesture-14/28 dataset.
	\end{abstract}
	

	\section{Introduction}
	This paper focuses on gesture analysis in the context of human behavioral biometrics. The paper describes a process of recognizing different gestures using only skeleton joint positions.   Behavior biometrics is being heavily utilized in the current developments of ``ambient'' intelligence and incorporating smart-home sensors for automated health monitoring and risk assessment, particularly in analyzing daily routines in smart homes \cite{dawadi2013automated}.
	
	The growing concern of elderly patients wanting to live independently is a major motivation for creating an automated system of analyzing behavior biometrics.  Due to the recent improvements in biometrics, better systems can be created for more accurate activity detection and prevention, specifically monitoring system that provides support when an emergency occurs.  A study by Lam et al. \cite{lam2015smartmind} investigated ambient monitoring and tracking  for Alzheimer patients performing daily activities.  An embedded sensing system proposed in \cite{lee2010embedded} was used to monitor how well an elderly patient performs an activity, as well as the potential to use it for determining the severity of the cognitive decline.  Using the activity-aware smart home behavior data, Alberdi et al. \cite{aramendi2018smart} created regression and classification models capable of predicting mobility, cognition and depression symptoms. \cite{minor2017learning} studies the problem of activity prediction and proposes using a combination of imitation learning and regression learners to address the problem.
	
	In recent years, deep learning techniques have greatly contributed to the increasing accuracy of existing biometric systems \cite{asadi2017survey}. In the area of human activity recognition (HAR), the development of deep learning methods overcome many existing pattern recognition challenges including: hand-crafted features, shallow features, a large number  of labeled data \cite{wang2018deep}. A 3D CNN proposed by \cite{molchanov2015hand} performs gesture recognition on both color and depth images and has achieved a classification rate of 77.5\% classification rate on the VIVA challenge dataset.  Chen et al. \cite{Chen2017} suggests using RNN to recognize different gestures by extracting global and finger motion features from a skeleton sequence.  Whereas \cite{Nunez2018} proposes to recognize dynamic hand gesture using only skeleton-based information with a combination of CNNs and long short term memory (LSTM) recurrent neural networks (RNN). An architecture proposed in \cite{feichtenhofer2016convolutional} shows the ability to perform a spatial-temporal fusion of video snippets used for action recognition.
	
	Gestures have been investigated as an alternative to the password for subject authentication.  Wu et al. \cite{wu2016two} proposed a two-stream CNN to learn the ``style'' of subjects performing different gestures which would allow the system to perform subject verification and identification. Another method proposed by Abate et al. \cite{abate2017implicitly} combines two biometrics, ear and arm gestures, collected using a smartphone to perform identity verification. A feasibility study on multi-touch gesture authentication system was reported in \cite{sae2014multitouch}. Study of subject authentication based on analysis of a swipe gesture,  Support Vector Machines, Gaussian Mixture Model, and fusion of both systems on four databases showed that horizontal swipes were more discriminatory \cite{fierrez2018benchmarking}.

	The objective of this study is to improve the performance of joint-based gesture recognition.  Traditional gesture recognition uses mainly static or video images for analysis.  Recent approaches incorporate other types of information such as depth, near-infrared, infrared, and time sequencing \cite{koppula2013learning}.  The proposed approach focuses on using a fusion classifier that is composed of two sub-networks.  Both sub-networks are types of deep learning architecture that was previously designed to handle sequence-to-sequence processing, such as language translation and natural language processing.  The first sub-network is based on an encoder-decoder and is called a transformer.  The second sub-network is a recurrent neural network (RNN) composed of the ordered neuron long-short-
	\newgeometry{left=0.75in, right=0.75in, top=0.75in, bottom=0.75in}
	\twocolumn[{\begin{figure}[H]
			\setlength{\linewidth}{\textwidth}
			\setlength{\hsize}{\textwidth}
			\centering
			\includegraphics[width=0.9\textwidth]{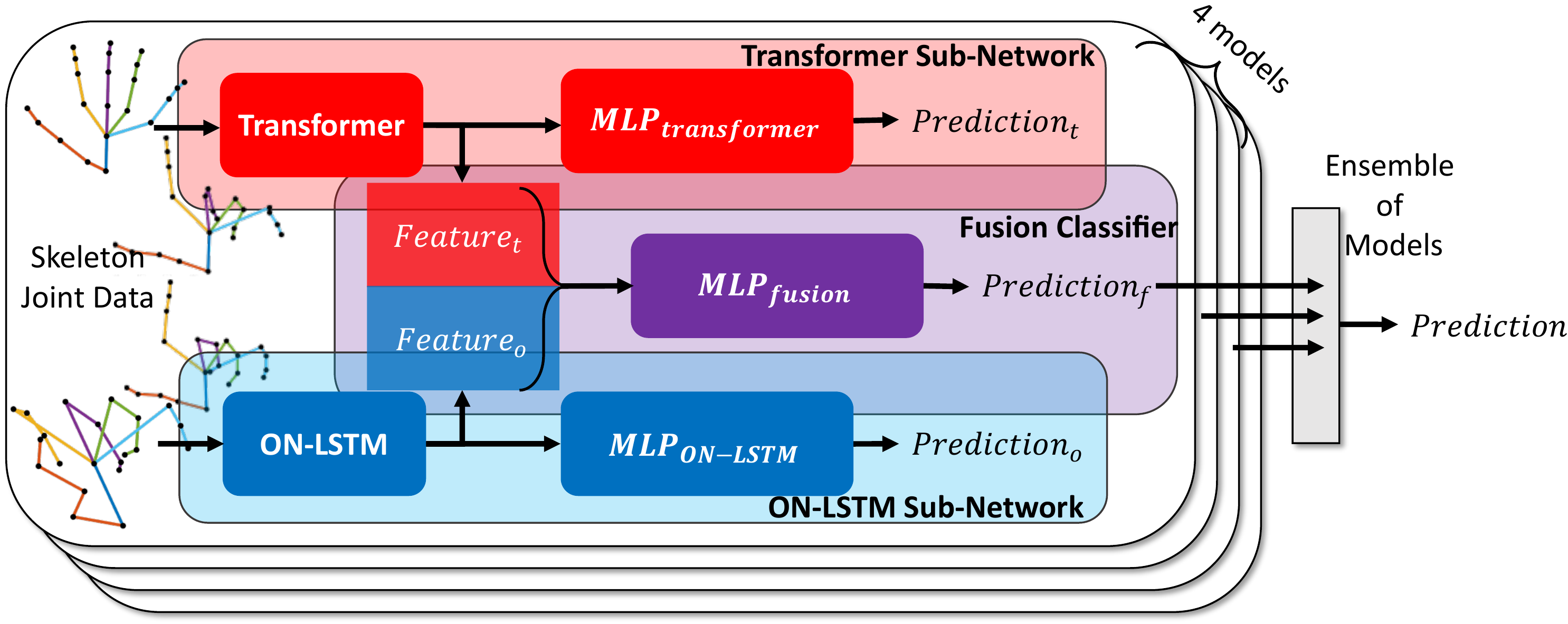} 
			\caption{The overall design of the proposed model.  The model consists of two independent sub-networks, Transformer and ON-LSTM, whose processed skeleton features are concatenated and classified to create a fusion classifier. The outputs of the fusion classifier are analyzed at four different instances and ensembled together to generate an ensemble prediction.}\label{fig:design}
	\end{figure}}]
	
	\noindent term-memory (LSTM) cell units.  Both sub-network can be independent and fully capable of performing gesture recognition. In this paper, we show that both of these networks can be fused to yield better performance through an approach known as knowledge distillation.
	
	The paper is structured as follows: overview of the proposed method is described in Section \ref{sec:overview}, design of experiments and experimental results are provided in Section \ref{sec:experiments}, and Section \ref{sec:conclusion} summarizes the findings.
	
	\section{Approach and Methodology}\label{sec:overview}
	In this paper, we propose a gesture recognition model composed of an ensemble of four models, where each model consists of four classifiers.  The combination of the classifiers is adopted from a process called Feature Fusion Learning proposed in \cite{kim2019feature}.  Two classifiers are independent sub-networks that are trained to perform gesture recognition, the logits of these classifiers are averaged to create an ensemble classifier, whereas the features of these classifiers are concatenated to create a fusion classifier.  Figure \ref{fig:design} illustrates the proposed model containing the three main classifiers.  The remaining classifier, a hidden ensemble classifier, does not directly output any prediction but is used purely to transfer its knowledge to the fusion classifier.
	
	The two independent classifiers are the two sub-networks, the transformer network and the ordered-neuron LSTM (ON-LSTM) based RNN.  The ensemble classifier is a single-layered network that averages the result of the logit inputs. Those units are named after logistic regression that estimates the parameters of a logistic model, using the unit of measurement for the log-odds scale called a logit, or logistic unit. 
	
	The fusion classifier is a simple three-layer multi-layer perceptron (MLP) that combines the feature inputs from the other networks.
	
\begin{figure}[!ht]
		\begin{center}
			\includegraphics[width=0.25\textwidth]{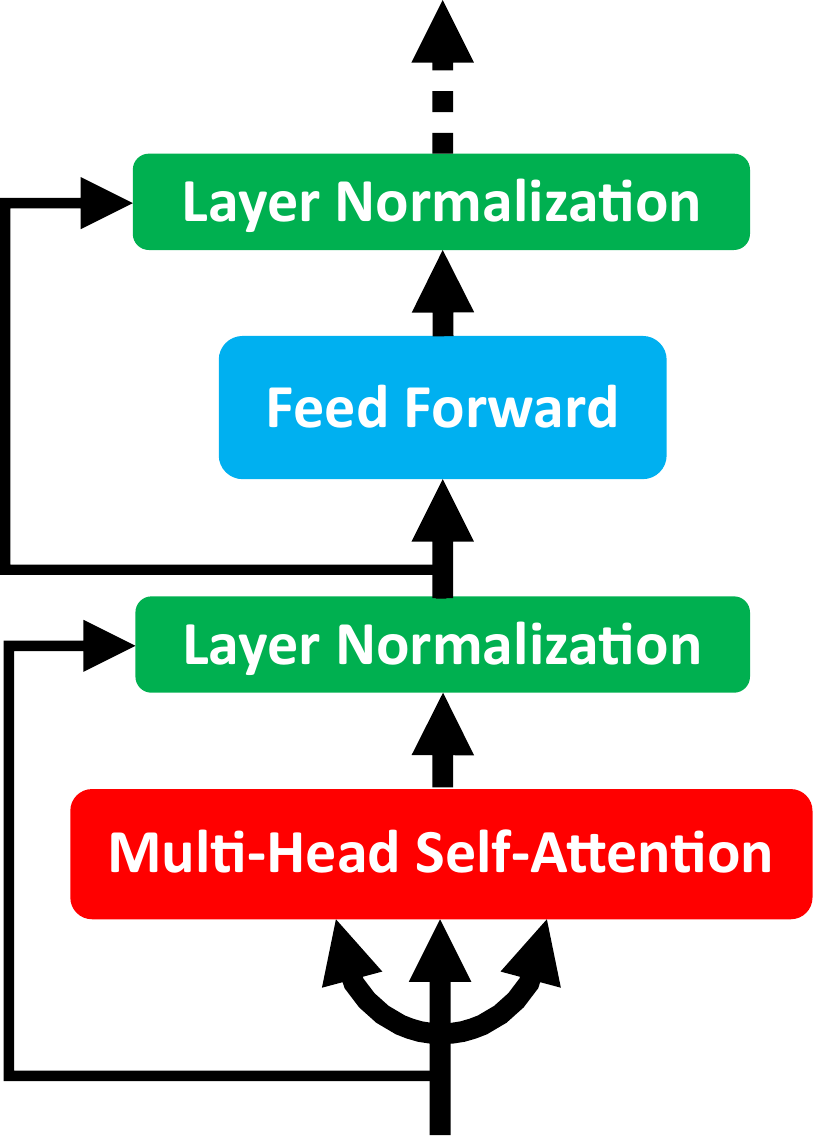} 
			\caption{Structure of a transformer block.}
			\label{fig:transformer}
		\end{center}
	\end{figure}

	The first sub-network uses a transformer structure which is a deep learning architecture that is designed to tackle sequence-to-sequence problems.  The Universal Transformer \cite{dehghani2019universal} is used in this paper as one of the sub-networks for analyzing the skeleton joints to recognize different gestures. Table \ref{tab:transformer} describes the transformer sub-network used in this paper.  A transformer block consists of several components including a multi-head self-attention, residual connection, layer normalization, feed-forward/transition function, residual connection and layer normalization as shown in Figure \ref{fig:transformer}.
	
	\begin{table}[!htb]
		\centering
		\caption{Transformer Sub-network}\label{tab:transformer}
		\begin{tabular}{lcccc}
			&		\multicolumn{3}{c}{Output Shape}	&	Parameters	\\
			\hline
			Input	&				64	&	$\times$	&	66*3	&	-	\\
			Transformer Block	&				64	&	$\times$	&	66		&	52.9K	\\
			Transformer Block	&				64	&	$\times$	&	66		&	52.9K	\\
			Transformer Block	&				64	&	$\times$	&	66		&	52.9K	\\
			Flatten				&				\multicolumn{3}{c}{		4224		}	&	-	\\
			\hline
			Fully-Connected		&	\multicolumn{3}{c}{		512		}	&	2.16M	\\
			Batch Normalization &  	\multicolumn{3}{c}{		512		}	&	2048	\\
			Mish Activation 	&  	\multicolumn{3}{c}{		512		}	&	-	\\
			Dropout (0.5)		&	\multicolumn{3}{c}{		512		}	&	-	\\
			
			Fully-Connected		&	\multicolumn{3}{c}{		512		}	&	263K	\\
			Batch Normalization &  	\multicolumn{3}{c}{		512		}	&	2048	\\
			Mish Activation 	& 	\multicolumn{3}{c}{		512		}	&	-	\\
			Dropout (0.5)		&	\multicolumn{3}{c}{		512		}	&	-	\\
			
			Fully-Connected		&	\multicolumn{3}{c}{		14		}	&	7182	\\
			Batch Normalization &  	\multicolumn{3}{c}{		14		}	&	56	\\
			Softmax Activation 	&  	\multicolumn{3}{c}{		14		}	&	-	\\
		\end{tabular}
	\end{table}

	The other sub-network consists of ON-LSTM cells which are proposed in \cite{shen2019ordered} as an initial step in integrating tree structures into an LSTM cell. The ON-LSTM shows promise in recognizing gestures as skeleton joints are interconnected. As well, it is suggested that normal LSTM may not be able to capture the relationship between different joints. The ON-LSTM cell is a modified version of the LSTM cell which introduces a master forget and master input gate.  The purpose of these gates is to control the update operations of the cell states at a high level.  This change governed by a new activation function, cumax, allows the cell to apply different updates to separate segments  \cite{shen2019ordered}.  
	
	Table \ref{tab:onlstm} describes the ON-LSTM sub-network used in this paper.
	
	\begin{table}[!htb]
		\centering
		\caption{ON-LSTM Sub-network}\label{tab:onlstm}
		\begin{tabular}{lcccc}
			&		\multicolumn{3}{c}{Output Shape}	&	Parameters	\\
			\hline
			Input	&				64	&	$\times$	&	66*3	&	-	\\
			ON-LSTM	&				64	&	$\times$	&	660		&	1.9M	\\
			ON-LSTM	&				\multicolumn{3}{c}{		660		}	&	3.5M	\\
			\hline
			Fully-Connected		&	\multicolumn{3}{c}{		512		}	&	338K	\\
			Batch Normalization &  	\multicolumn{3}{c}{		512		}	&	2048	\\
			Mish Activation 	&  	\multicolumn{3}{c}{		512		}	&	-	\\
			Dropout (0.5)		&	\multicolumn{3}{c}{		512		}	&	-	\\
			
			Fully-Connected		&	\multicolumn{3}{c}{		512		}	&	263K	\\
			Batch Normalization &  	\multicolumn{3}{c}{		512		}	&	2048	\\
			Mish Activation 	& 	\multicolumn{3}{c}{		512		}	&	-	\\
			Dropout (0.5)		&	\multicolumn{3}{c}{		512		}	&	-	\\
			
			Fully-Connected		&	\multicolumn{3}{c}{		14		}	&	7182	\\
			Batch Normalization &  	\multicolumn{3}{c}{		14		}	&	56	\\
			Softmax Activation 	&  	\multicolumn{3}{c}{		14		}	&	-	\\
		\end{tabular}
	\end{table}
	
	\subsection{Mish Activation}
	
	The first technique applied to improve the performance of the model is the replacement of the rectified linear activation units (ReLU) with Mish.  Mish is non-monotonic and is continuously differentiable which may resolve gradients problem associated with rectified linear units, specifically in the case for input values less than equal to 0 \cite{misra2019mish}. It was shown to provide better promise than Swish and ReLU \cite{misra2019mish}. Swish is used in Bidirectional Encoder Representations from Transformers (BERT) which has reached state-of-the-art performance in natural language processing.  Mish is defined as follows:
	\begin{equation} \label{eq:mish}
	f(x)=x\cdot tanh(ln(1+e^x))
	\end{equation}
	where $x$ is the input value, $tanh$ is the hyperbolic tangent function, and $ln$ is the natural logarithmic function.
	\subsection{Knowledge Distillation}
	
	Another technique involves knowledge distillation, specifically the transfer of the dark knowledge from the ensemble classifier to the fusion classifier.  Dark knowledge refers to the hidden information learned by the models and can be revealed by calculating the softened probability based on a temperature $T$, as defined in Equation \ref{eq:sp} \cite{hinton2015distilling}. When $T=1$, the resulting probability is the same as the result of a softmax function.  The dark knowledge from the fusion classifier is then also distilled to the two sub-networks. 
	
	\begin{equation} \label{eq:sp}
	\sigma_{i,m}= \frac{\exp{(Logit^i_m/T)}}{\sum_{j}^{N}\exp{(Logit^j_m/T)}}
	\end{equation}
	where $N$ is the total number of classes, $T$ is the temperature parameter, $Logit^i_m$ is the $i^{th}$ class's logit output from $m$ network.
	
	In this paper, we apply two knowledge distillation proposed in \cite{kim2019feature}: ensemble knowledge distillation and fusion knowledge distillation.  Ensemble knowledge distillation (EKD) transfers knowledge from the ensemble classifier to the fusion classifier. The fusion knowledge distillation (FKD)  transfers the new knowledge of the ensemble classifier to the individual sub-networks.  
	
\begin{figure*}[!ht]
		\begin{center}
			\includegraphics[width=0.9\textwidth]{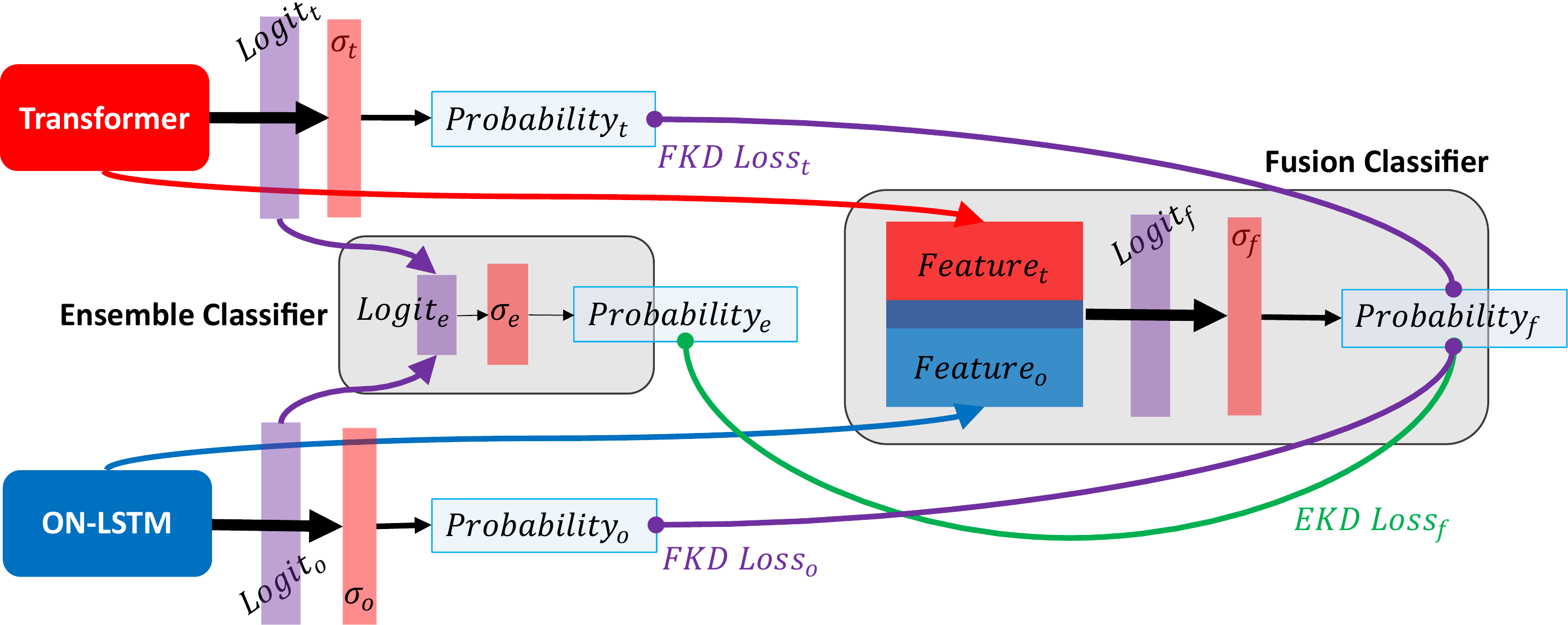} 
			\caption{Knowledge Distillation between (1) the fusion classifier and the ensemble classifier (EKD$_f$) and (2) the fusion classifier and the transformer/ON-LSTM classifier (FKD$_t$ and FKD$_o$) .  Logit$_e$ is calculated as the average between Logit$_t$ and Logit$_o$.}
			\label{fig:distillation}
		\end{center}
	\end{figure*}

	Figure \ref{fig:distillation} illustrates the two types of knowledge distillation used to transfer knowledge between the classifiers.
	
	EKD loss is defined as the Kullback-Leibler divergence between the softened distribution of the ensemble classifier and the softened distribution of the fusion classifier.   FKD loss is defined as the Kullback-Leibler divergence between the softened distribution of the fusion classifier and the softened distribution of the individual networks.
	
	The loss for the individual sub-network is the combination of the cross-entropy (CE) loss and the FKD loss defined as follows:
	\begin{equation} \label{eq:loss1}
	\begin{split}
	L_{m}= \underbrace{\sum_{i}^{N}\sigma_{i,f} \log(\frac{\sigma_{i,f}}{\sigma_{i,m}})}_\text{FKD Loss} \underbrace{- \sum_{i}^{N}y_i \log(\sigma_{i,m})}_\text{CE Loss}
	\end{split}
	\end{equation}

	The loss for the fusion classifier is the combination of the original CE loss accompanied with the EKD loss defined in Equation \ref{eq:loss2} as follows:
	\begin{equation} \label{eq:loss2}
	\begin{split}
	L_{f}= \underbrace{\sum_{i}^{N}\sigma_{i,e} \log(\frac{\sigma_{i,e}}{\sigma_{i,f}})}_\text{EKD Loss} \underbrace{- \sum_{i}^{N}y_i \log(\sigma_{i,f})}_\text{CE Loss}
	\end{split}
	\end{equation}
	where $L_m$ is the loss for the $m$ sub-network, $L_f$ is the loss for the fusion classifier, $y_i$ is the truth label for the $i^{th}$ index in a one-hot-encoded label, and $\sigma_{i,m}$, $\sigma_{i,f}$, and $\sigma_{i,e}$ represents the soften probabilities for the $m$ sub-network, fusion classifier, and ensemble classifier, respectively.

	\subsection{AdamWR and an Ensemble of Models}
	
	In this paper, an ensemble of models is used to further boost the performance of the recognition accuracy. Specifically, the outputs of the four models are combined to yield a more confident prediction.
	
	To achieve the creation of multiple models without re-training a model, a technique that combines a cyclical learning rate and warm restart is used.  The cyclical learning rate is a technique that cyclically varies a learning rate between two bounds.  This technique shows an improvement over monotonically decreasing a learning rate and relieves the necessary processing of fine-tuning the learning rate \cite{smith2017cyclical}.  Cosine annealing, another technique that performs a warm restart of the training process with a learning rate adjusting to a cosine wave \cite{loshchilov2017sgdr}.  By using cosine annealing, the model parameters can be saved at the end of each cycle.  
	
	Built upon using cyclical learning rates, it is possible to save the model parameters at the end of each cycle \cite{huang2017snapshot}.  At the end of the training, $K$ models are saved after $K$ cycles, and through the ensemble fusion of the M models, the system can obtain better performance.  Since each model is optimized for a specific cycle of learning rate and that each cycle is capable of dislodging from a local minimum, the fusion of multiple models contributes to different decisions.
	
	Lastly, a weight decay is applied to the Adam optimizer as a way to incorporate better regularization in the model \cite{loshchilov2019decoupled}.  Two common ways of regularization are used in neural networks, specifically L1 and L2 regularization.  The application of these regularization reduces bias and variance, respectively.  However, the incorporation of L2 regularization is different for adaptive gradient methods, such as Adam.  In \cite{loshchilov2019decoupled}, a technique to decouple weight decay from the gradient is proposed.

	\subsection{Data Augmentation}
\begin{figure}[!ht]
		\begin{center}
			\includegraphics[width=0.48\textwidth,interpolate]{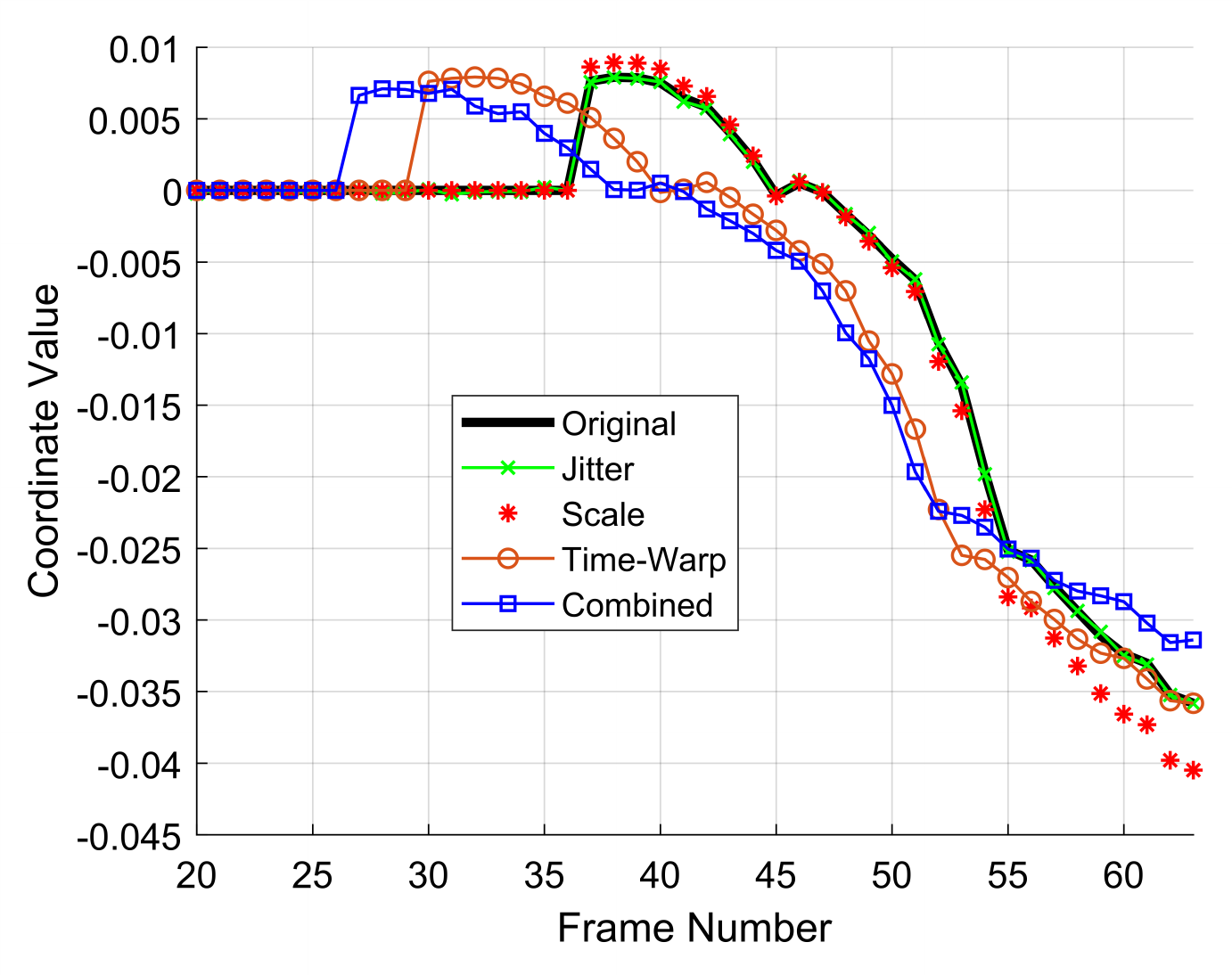} 
			\caption{Data augmentation on the x-coordinate value of wrist joint.}
			\label{fig:augment}
		\end{center}
	\end{figure}

	Due to the nature of deep neural networks, it is often beneficial to have a large training dataset.  In this paper, we apply different augmentation techniques on the DHG-14/28 dataset to greatly increase the number of training samples.  The creation of synthetic time-series data, specifically for skeleton joints, is different from conventional augmentation of images.  Four augmentation technique is described in \cite{rashid2019times}, which is used to augment inertial measurement unit data.  In this paper, we adopt three of the described technique, namely jittering, scaling and time-warping, and generate a fourth technique by combining the previous three methods.  Figure \ref{fig:augment} shows an example of data augmentation applied on the x-coordinates of the wrist joint for subject 1 while performing the grabbing gesture.
	
	\subsubsection{Jittering}
	Simulating sensor noise by applying Gaussian noise to the x, y, and z-coordinates of the joint.  The noise is a Gaussian distribution with a mean of 0 and a variance calculated based on the coordinate axes. 
	\begin{equation}
	X=X+N(\mu_x=0, Var(X))
	\end{equation}
	
	\subsubsection{Scaling}
	Modify the magnitude of the data by applying a random scalar value between 0.75 and 1.25 to the x, y, and z-coordinates of the joints.
	\begin{equation}
	X=X\cdot Random(0.75,1.25)
	\end{equation}
	
	\subsubsection{Time-warping}
	Adjusting the overall duration of the time-series data. A time-series with length $T$ is interpolated to a series with length $T*V$ where $V$ is a random value between 0.5 and 2.  A $V>1$ increases the length of the time-series while a $V<1$ shortens the series.
	
	\section{Experiments}\label{sec:experiments}
	
	The proposd model was trained for 4 cycles with the decoupled weight decay Adam optimizer with warm restart (AdamWR).  For each subsequent cycle, the number of epochs is raised by a factor of 1.5.  In total there are 10+15+23+35= 83 epochs.  For each cycle, a cosine-based cyclical learning rate with a starting learning rate $\alpha=0.001$, $\beta_1=0.9$, $\beta_2=0.999$, and weight decay of 0.001 is used. For the transformer blocks, 11 heads were selected for self attention.  A temperature $T=3$ is used to compute the softened probability for knowledge distillation.   At each cycle, a snapshot of the model is saved for an ensemble of models.  In addition, a time-step of 64 with a batch size of 512 is used for the training of the model.  The input data for the model are the skeleton joint positions encoded as world coordinates (22 joints by x, y, and z-coordinates).  The output of the networks is the predictions from the four classifiers, fusion, transformer, and ON-LSTM.
	
	For evaluating the performance of the model, we choose to use a specific type of $K$-fold cross-validation, that divides each fold based on subjects.  This leave-one subject-out cross-validation (LOOCV) is further described in \cite{Smedt2016} and is one of the main methods for comparing different gesture recognition algorithms.  Given the DHG-14/28 dataset contains 20 subjects, the dataset is augmented by a factor of 40 and split into 20 folds, a fold for each subject.  When performing LOOCV, the 19 augmented-folds are used for training while the remaining fold is used for validation. A third set, the testing set, consists of the original un-augmented validation fold is used for calculating the final accuracy. Performance of the system is evaluated in terms of the accuracy of action recognition, defined as follows: 
	\[
	\texttt{Acc} = \frac{TP+TN}{TP+FP+TN+FN}
	\]
	where $TP$, $TN$, $FP$, and $FN$ represent the number of true positives, true negatives, false positives and false negatives, respectively. Accuracy reflects the system's ability to accept genuine actions while rejecting imposter actions.  Due to using the LOOCV method, the overall reported accuracy is the result of averaging the accuracies for each model trained on different subjects.  For example, given twenty subjects, there are twenty models; each model is trained on nineteen subjects and is tested on the remaining subject, resulting in twenty estimated accuracy values. The overall accuracy is the average of these twenty values.
	
	\subsection{Datasets}
\begin{figure}[!ht]
		\begin{center}
			\begin{tabular}{cccc} 
				\includegraphics[width=0.0736\textwidth]{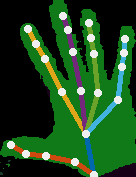} &
				\includegraphics[width=0.10\textwidth]{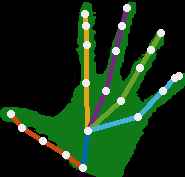} &
				\includegraphics[width=0.1154\textwidth]{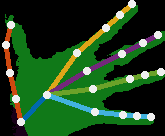} &
				\includegraphics[width=0.0952\textwidth]{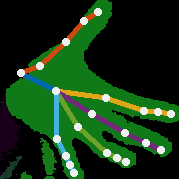} \\
				$(a)$ & $(b)$ & $(c)$ & $(d)$
			\end{tabular}
			\caption{Illustrations of the rotate clockwise gesture from the DHG-14/28 \cite{Smedt2016}.  $(a)$ Frame 21, $(b)$ Frame 30,  $(c)$ Frame 36,  $(d)$ Frame 57.
			}\label{fig:activities}
		\end{center}
	\end{figure}

	The dynamic hand gesture 14/28 (DHG-14/28) \cite{Smedt2016} was chosen as the database, and this is one of the few databases containing data collected using a depth camera (Intel RealSense F200) sensor. Both depth and skeleton information for various hand gestures is available.  In the DHG-14/28 \cite{Smedt2016} dataset, 20 unique individuals are performing 5 iterations of 14 gestures using two types of finger configurations, thus forming 28 sets of gestures, to a total of 2800 sequences. The depth information is saved in the form of images with a resolution of 480x640 in 16-bits. The skeleton information contains 22 joint locations of a hand described in both 2D and 3D coordinates saved in  44x1 and 66x1 vector format, respectively.  Figure \ref{fig:activities} illustrates an example of the rotate clockwise gesture where the skeleton joints are overlayed on top of the depth images.
	
	For the DHG-14/28 dataset, each gesture is individually classified into two main categories: fine-grained and coarse-grained gestures.  Table \ref{tab:gestures} provides a list of all the gestures and the corresponding grain categories.
	\begin{table}[!htb]
		\centering
		\caption{List of Gestures in the DHG-14/28 dataset}\label{tab:gestures}
		\begin{footnotesize}
			\begin{tabular}{l|l|l}
				Gesture & Grain & Tag Name\\
				\hline
				Grab & Fine & G\\
				Tap & Coarse & T\\
				Expand & Fine & E\\
				Pinch & Fine & P\\
				Rotation Clockwise & Fine & R-CW\\
				Rotation Counter-clockwise & Fine & R-CCW\\
				Swipe Right & Coarse & S-R\\
				Swipe Left & Coarse & S-L\\
				Swipe Up & Coarse & S-U\\
				Swipe Down & Coarse & S-D\\
				Swipe X & Coarse & S-X\\
				Swipe V & Coarse & S-V\\
				Swipe + & Coarse & S-+\\
				Shake & Coarse & Sh\\
			\end{tabular}
		\end{footnotesize}
	\end{table}
	
	\subsection{Experimental results}
	In our experiment, we evaluated the performance of our proposed model using the DHG-14/28 dataset.  Table \ref{table:accuracy1}  provides the performance of the different classifiers within the proposed model evaluated at different cycles.  Due to snap-shotting the model according to the learning rate associated with cosine annealing, four different models can be evaluated.  For example, the first cycle is associated with the model snap-shotted after 10 epochs.  It can be seen that the performance between the different cycles is relatively similar, but when combined in an ensemble together, it can lead to a performance increase. This indicates that the prediction from each cycle is slightly different and that new information can be retrieved by observing the discrepancy in prediction.

	\begin{table}[!htb]
		\caption{Accuracy (\%) performance at each cycle on the DHG-14/28 dataset}
		\begin{center}
			\begin{tabular}{c|c|ccc}
				Cycle	& Network &  Fine &  Coarse & Both \\
				\hline
				\multirow{3}{*}{1}&	Transformer	&	80.50&84.94	&83.36	\\
				&	ON-LSTM	&76.70	&86.28	&	82.86\\
				&	Fusion						&76.70	&83.39	&81.00	\\
				\hdashline
				\multirow{3}{*}{2}&	Transformer	&76.70	&81.83	&80.00	\\
				&	ON-LSTM	&71.40	&84.17	&79.61	\\
				&	Fusion						&78.00	&86.33	&83.36	\\
				\hdashline
				\multirow{3}{*}{3}&	Transformer	&79.90	&84.94	&83.14	\\
				&	ON-LSTM	&76.30	&86.83	&83.07	\\
				&	Fusion						&77.80	&87.33	&83.93	\\
				\hdashline					
				\multirow{3}{*}{4}&	Transformer	&	81.60&86.39	&84.68	\\
				&	ON-LSTM	&77.40	&87.00	&	83.57\\
				&	Fusion						&78.00	& 86.72	& 83.61	\\
				\hline
				\multirow{3}{*}{Ensemble}&	Transformer	&	\textbf{81.70}&86.72&84.93	\\
				&	ON-LSTM	&78.00	&87.39	&	84.04\\
				&	Fusion						&	81.20	&	\textbf{88.83}	&	\textbf{86.11}	\\
			\end{tabular}
		\end{center}
		\label{table:accuracy1}
	\end{table}
	
	Table \ref{table:accuracy2} provides a list of gesture recognition methods applied to the DHG-14/28 dataset.  Our proposed ensemble model yields similar but slightly improved accuracy compared to the other methods.  The main contribution to the higher accuracy is a significant increase in recognizing ``fine''-grained gestures but at a slight cost of decreasing ``coarse''-grained gestures.
	
	\begin{table}[!htb]
		\caption{Comparison of accuracy (\%) with various methods on the DHG-14/28 dataset}
		\begin{center}
			\begin{tabular}{c|ccc}
				Method &  Fine &  Coarse & Both \\
				\hline
				SOCJ+HoHD+HoHR \cite{de2016skeleton}	&	73.60	&	88.33	&	83.07	\\
				NIUKF-LSTM	\cite{ma2018hand}	&	-	&	-	&	84.92	\\
				SL-fusion-Average \cite{laiCNN2018} &	76.00&	90.72&	85.46	\\
				CNN+LSTM \cite{Nunez2018}		&	78.00	&	89.80	&	85.60	\\
				MFA-Net		\cite{chen2019mfa}	&	75.60	&	\textbf{91.39}	&	85.75	\\
				Ensemble of Models (Ours)			&	\textbf{81.20}	&	88.83	&	\textbf{86.11}	\\
			\end{tabular}
		\end{center}
		\label{table:accuracy2}
	\end{table}

\begin{figure}[!ht]
		\begin{center}
			\includegraphics[width=0.48\textwidth]{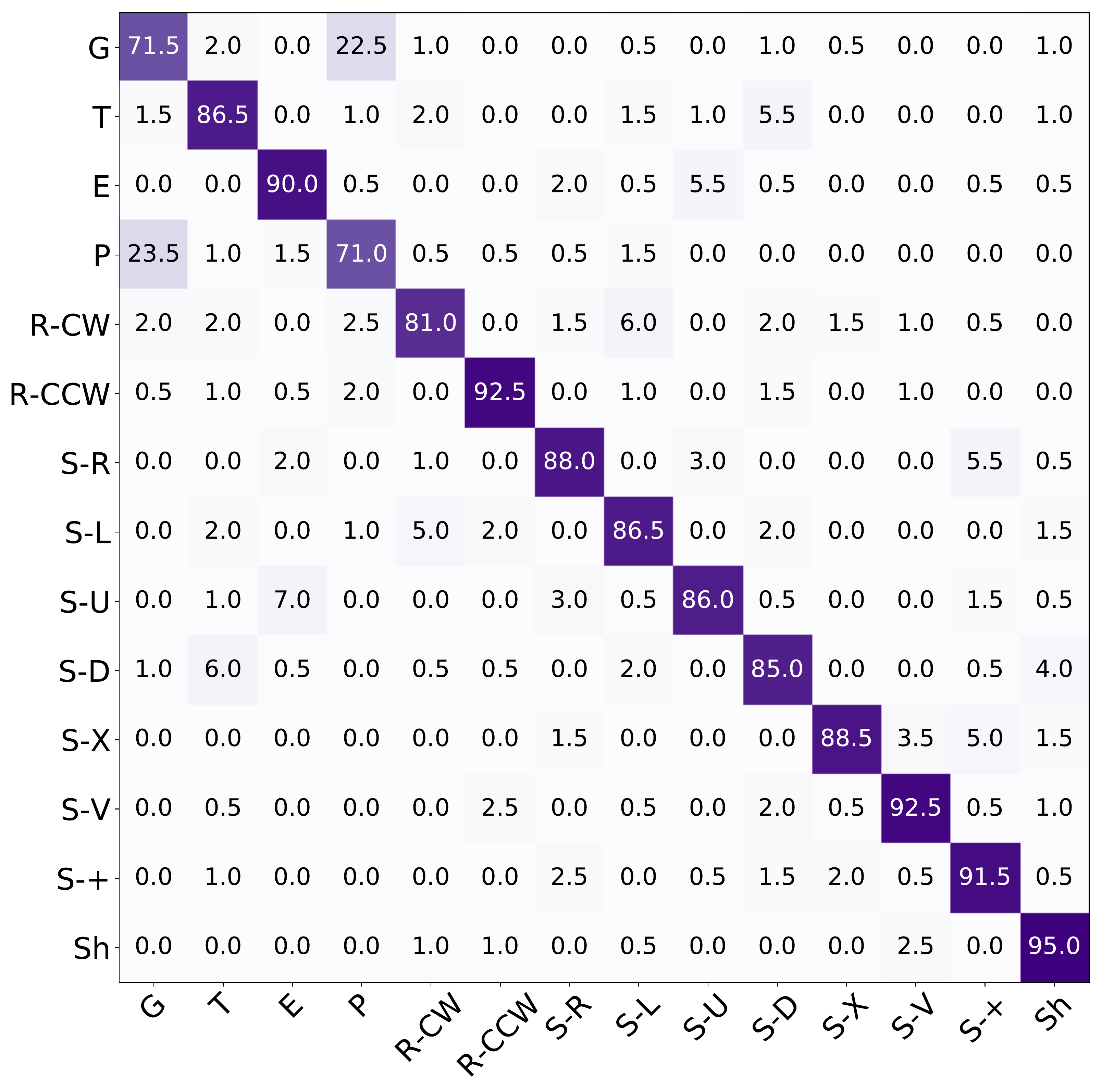} 
			\caption{The confusion matrix showing accuracies of gesture recognition on 14 gestures when using the proposed method of an ensemble of four models.}
			\label{fig:cm}
		\end{center}
	\end{figure}

	To further evaluate the model, a confusion matrix was generated to capture the distribution of the different gestures being recognized.  Figure \ref{fig:cm} illustrates the performance of the proposed ensemble model when recognizing the 14 gestures.  Each row represents the sample gesture and each column indicates the model's prediction.  For example, the last row of Figure \ref{fig:cm} refers to the Shake gesture and that 95\% of these samples are correctly recognized by the model while the remaining 5\% of the samples are split amongst R-CW, R-CCW, and S-V gestures.  Figure \ref{fig:cm} reveals that one of the main reasons for the increase in performance for ``fine''-grained gestures is due to the model's ability to distinguish between the grabbing and pinching gestures.  It was mentioned previously in \cite{Chen2017} and \cite{Smedt2016} that the grabbing and pinching gesture is very similar and that the main difference lies in the amplitude of the action.
	
	\section{Conclusion}\label{sec:conclusion}
	This paper proposes an ensemble of knowledge-sharing models designed to improve the performance of gesture recognition.  Each model consists of multiple sub-networks that are fully capable of independently performing gesture recognition. However, since each network is developed using a different architecture, namely transformer and ON-LSTM, the features each network extracts are different.  We propose to fuse the results from each of the sub-networks to create a more robust model.  By applying knowledge distillation, the proposed model benefits from the knowledge learned by each of the sub-networks.  The proposed model yields an accuracy of 86.11\% using only skeleton joint positions.  Also, the model reveals that using knowledge distillation to fuse features helps in identifying hard-to-recognize gestures.  This knowledge sharing property helps optimize the sub-networks which, therefore, learn how to distinguish between troublesome classes.

	\section*{Acknowledgment}
	This work was partially supported by Natural Sciences and
	Engineering Research Council of Canada through Discovery
	Grant ``Biometric Intelligent Interfaces'', and by the Department of National Defense’s Innovation for Defense Excellence and Security (IDEaS) program, Canada.
	
	{\small
		\bibliographystyle{IEEEtran}
		\bibliography{gest}

\begin{thebibliography}{10}
\providecommand{\url}[1]{#1}
\csname url@samestyle\endcsname
\providecommand{\newblock}{\relax}
\providecommand{\bibinfo}[2]{#2}
\providecommand{\BIBentrySTDinterwordspacing}{\spaceskip=0pt\relax}
\providecommand{\BIBentryALTinterwordstretchfactor}{4}
\providecommand{\BIBentryALTinterwordspacing}{\spaceskip=\fontdimen2\font plus
\BIBentryALTinterwordstretchfactor\fontdimen3\font minus
  \fontdimen4\font\relax}
\providecommand{\BIBforeignlanguage}[2]{{%
\expandafter\ifx\csname l@#1\endcsname\relax
\typeout{** WARNING: IEEEtran.bst: No hyphenation pattern has been}%
\typeout{** loaded for the language `#1'. Using the pattern for}%
\typeout{** the default language instead.}%
\else
\language=\csname l@#1\endcsname
\fi
#2}}
\providecommand{\BIBdecl}{\relax}
\BIBdecl

\bibitem{dawadi2013automated}
P.~N. Dawadi, D.~J. Cook, and M.~Schmitter-Edgecombe, ``Automated cognitive
  health assessment using smart home monitoring of complex tasks,'' \emph{IEEE
  Trans. on Systems, Man, and Cybernetics: Systems}, vol.~43, no.~6, pp.
  1302--1313, 2013.

\bibitem{lam2015smartmind}
K.-Y. Lam, N.~W.-H. Tsang, S.~Han, J.~K.-Y. Ng, S.-W. Tam, and A.~Nath,
  ``Smartmind: Activity tracking and monitoring for patients with alzheimer's
  disease,'' in \emph{IEEE Int. Conf. on Advanced Information Networking and
  Applications}, 2015, pp. 453--460.

\bibitem{lee2010embedded}
M.~L. Lee and A.~K. Dey, ``Embedded assessment of aging adults: a concept
  validation with stakeholders,'' in \emph{Int. Conf. on Pervasive Computing
  Technologies for Healthcare}, March 2010, pp. 1--8.

\bibitem{aramendi2018smart}
A.~A. Aramendi, A.~Weakley, M.~Schmitter-Edgecombe, D.~J. Cook, A.~A. Goenaga,
  A.~Basarab, and M.~B. Carrasco, ``Smart home-based prediction of multidomain
  symptoms related to alzheimer's disease,'' \emph{IEEE Journal of Biomedical
  and Health Informatics}, vol.~22, no.~6, pp. 1720--1731, 2018.

\bibitem{minor2017learning}
B.~D. Minor, J.~R. Doppa, and D.~J. Cook, ``Learning activity predictors from
  sensor data: Algorithms, evaluation, and applications,'' \emph{IEEE Trans. on
  Knowledge and Data Engineering}, vol.~29, no.~12, pp. 2744--2757, December
  2017.

\bibitem{asadi2017survey}
M.~Asadi-Aghbolaghi, A.~Clapes, M.~Bellantonio, H.~J. Escalante,
  V.~Ponce-L{\'o}pez, X.~Bar{\'o}, I.~Guyon, S.~Kasaei, and S.~Escalera, ``A
  survey on deep learning based approaches for action and gesture recognition
  in image sequences,'' in \emph{IEEE Int. Conf. on Automatic Face \& Gesture
  Recognition}, 2017, pp. 476--483.

\bibitem{wang2018deep}
J.~Wang, Y.~Chen, S.~Hao, X.~Peng, and L.~Hu, ``Deep learning for sensor-based
  activity recognition: A survey,'' \emph{Pattern Recognition Letters}, 2018.

\bibitem{molchanov2015hand}
P.~Molchanov, S.~Gupta, K.~Kim, and J.~Kautz, ``Hand gesture recognition with
  3d convolutional neural networks,'' in \emph{IEEE Conf. on Computer Vision
  and Pattern Recognition Workshops}, June 2015, pp. 1--7.

\bibitem{Chen2017}
X.~Chen, H.~Guo, G.~Wang, and L.~Zhang, ``Motion feature augmented recurrent
  neural network for skeleton-based dynamic hand gesture recognition,'' in
  \emph{IEEE Int. Conf. on Image Processing}, September 2017, pp. 2881--2885.

\bibitem{Nunez2018}
J.~C. N{\'u}{\~n}ez, R.~Cabido, J.~J. Pantrigo, A.~S. Montemayor, and J.~F.
  V{\'e}lez, ``Convolutional neural networks and long short-term memory for
  skeleton-based human activity and hand gesture recognition,'' \emph{Pattern
  Recognition}, vol.~76, pp. 80 -- 94, 2018.

\bibitem{feichtenhofer2016convolutional}
C.~Feichtenhofer, A.~Pinz, and A.~Zisserman, ``Convolutional two-stream network
  fusion for video action recognition,'' in \emph{IEEE Conf. on Computer Vision
  and Pattern Recognition}, 2016, pp. 1933--1941.

\bibitem{wu2016two}
J.~Wu, P.~Ishwar, and J.~Konrad, ``Two-stream cnns for gesture-based
  verification and identification: Learning user style,'' in \emph{IEEE Conf.
  on Computer Vision and Pattern Recognition Workshops}, 2016, pp. 42--50.

\bibitem{abate2017implicitly}
A.~F. Abate, M.~Nappi, and S.~Ricciardi, ``I-am: implicitly authenticate me
  person authentication on mobile devices through ear shape and arm gesture,''
  \emph{IEEE Trans. on Systems, Man, and Cybernetics: Systems}, no.~99, pp.
  1--13, 2017.

\bibitem{sae2014multitouch}
N.~Sae-Bae, N.~Memon, K.~Isbister, and K.~Ahmed, ``Multitouch gesture-based
  authentication,'' \emph{IEEE Trans. on information forensics and security},
  vol.~9, no.~4, pp. 568--582, 2014.

\bibitem{fierrez2018benchmarking}
J.~Fierrez, A.~Pozo, M.~Martinez-Diaz, J.~Galbally, and A.~Morales,
  ``Benchmarking touchscreen biometrics for mobile authentication,'' \emph{IEEE
  Trans. on Information Forensics and Security}, vol.~13, no.~11, pp.
  2720--2733, 2018.

\bibitem{koppula2013learning}
H.~S. Koppula, R.~Gupta, and A.~Saxena, ``Learning human activities and object
  affordances from rgb-d videos,'' \emph{Int. Journal of Robotics Research},
  vol.~32, no.~8, pp. 951--970, 2013.

\bibitem{kim2019feature}
J.~Kim, M.~Hyun, I.~Chung, and N.~Kwak, ``Feature fusion for online mutual
  knowledge distillation,'' \emph{arXiv preprint arXiv:1904.09058}, 2019.

\bibitem{dehghani2019universal}
M.~Dehghani, S.~Gouws, O.~Vinyals, J.~Uszkoreit, and {\L}.~Kaiser, ``Universal
  transformers,'' \emph{Int. Conf. on Learning Representations}, 2019.

\bibitem{shen2019ordered}
Y.~Shen, S.~Tan, A.~Sordoni, and A.~Courville, ``Ordered neurons: Integrating
  tree structures into recurrent neural networks,'' in \emph{Int. Conf. on
  Learning Representations}, 2019.

\bibitem{misra2019mish}
D.~Misra, ``Mish: A self regularized non-monotonic neural activation
  function,'' \emph{arXiv preprint arXiv:1908.08681}, 2019.

\bibitem{hinton2015distilling}
G.~Hinton, O.~Vinyals, and J.~Dean, ``Distilling the knowledge in a neural
  network,'' \emph{arXiv preprint arXiv:1503.02531}, 2015.

\bibitem{smith2017cyclical}
L.~N. Smith, ``Cyclical learning rates for training neural networks,'' in
  \emph{IEEE Winter Conf. on Applications of Computer Vision}, 2017, pp.
  464--472.

\bibitem{loshchilov2017sgdr}
I.~Loshchilov and F.~Hutter, ``Sgdr: Stochastic gradient descent with warm
  restarts,'' in \emph{Int. Conf. on Learning Representations}, 2017.

\bibitem{huang2017snapshot}
G.~Huang, Y.~Li, G.~Pleiss, Z.~Liu, J.~E. Hopcroft, and K.~Q. Weinberger,
  ``Snapshot ensembles: Train 1, get m for free,'' in \emph{Int. Conf. on
  Learning Representations}, 2017.

\bibitem{loshchilov2019decoupled}
I.~Loshchilov and F.~Hutter, ``Decoupled weight decay regularization,'' in
  \emph{Int. Conf. on Learning Representations}, 2019.

\bibitem{rashid2019times}
K.~M. Rashid and J.~Louis, ``Times-series data augmentation and deep learning
  for construction equipment activity recognition,'' \emph{Advanced Engineering
  Informatics}, vol.~42, p. 100944, 2019.

\bibitem{Smedt2016}
Q.~{De Smedt}, H.~Wannous, and J.~P. Vandeborre, ``Skeleton-based dynamic hand
  gesture recognition,'' in \emph{IEEE Int. Conf. on Computer Vision and
  Pattern Recognition Workshops}, June 2016, pp. 1206--1214.

\bibitem{de2016skeleton}
Q.~De~Smedt, H.~Wannous, and J.-P. Vandeborre, ``Skeleton-based dynamic hand
  gesture recognition,'' in \emph{IEEE Conf. on Computer Vision and Pattern
  Recognition Workshops}, 2016, pp. 1--9.

\bibitem{ma2018hand}
C.~Ma, A.~Wang, G.~Chen, and C.~Xu, ``Hand joints-based gesture recognition for
  noisy dataset using nested interval unscented kalman filter with lstm
  network,'' \emph{The visual computer}, vol.~34, no. 6-8, pp. 1053--1063,
  2018.

\bibitem{laiCNN2018}
K.~Lai and S.~N. Yanushkevich, ``{CNN+RNN Depth and Skeleton based Dyanamic
  Hand Gesture Recognition},'' in \emph{Int. Conf. on Pattern Recognition},
  August 2018, pp. 1--6.

\bibitem{chen2019mfa}
X.~Chen, G.~Wang, H.~Guo, C.~Zhang, H.~Wang, and L.~Zhang, ``Mfa-net: Motion
  feature augmented network for dynamic hand gesture recognition from skeletal
  data,'' \emph{Sensors}, vol.~19, no.~2, p. 239, 2019.

\end{thebibliography}
	}

\end{document}